\documentclass[runningheads]{llncs}
\usepackage{makeidx}
\usepackage{graphicx}
\pdfoutput=1
\usepackage[pdftex]{}
\usepackage{amsmath,amssymb} 
\usepackage{color}
\usepackage{listings}
\usepackage{url}
\usepackage{color, colortbl}
\definecolor{LRed}{rgb}{1,.8,.8}
\definecolor{MRed}{rgb}{1,.6,.6}
\definecolor{HRed}{rgb}{1,.2,.2}
\usepackage[pagebackref=true,breaklinks=true,letterpaper=true,colorlinks,bookmarks=false]{hyperref}

\newcommand\DAGMreviewversion{
	\usepackage{lineno}
	\usepackage{color}
	\renewcommand\thelinenumber{\color[rgb]{0.2,0.5,0.8}\normalfont\sffamily\scriptsize\arabic{linenumber}\color[rgb]{0,0,0}}
	\renewcommand\makeLineNumber {\hss\thelinenumber\ \hspace{6mm} \rlap{\hskip\textwidth\ \hspace{6.5mm}\thelinenumber}} 
}

\DAGMreviewversion

\begin{document}
	\pagestyle{headings}
	\mainmatter

	\title{Automatic 3D Liver Segmentation Using Sparse Representation of Global and Local Image Information via Level Set Formulation}

	\titlerunning{ Automatic 3D Liver Segmentation}
	\authorrunning{S. Al-Shaikhli et al.}
\author{Saif Dawood Salman Al-Shaikhli\inst{1} Michael Ying Yang\inst{2} \and
Bodo Rosenhahn\inst{1}}

\institute{tnt Institute for Information Processing / Leibniz University Hannover, Germany,
\and
Computer Vision Lab. / TU Dresden, Germany}

	\maketitle

	\begin{abstract}
In this paper, a novel framework for automated liver segmentation via a level set formulation is presented. A sparse representation of both global (region-based) and local (voxel-wise) image information is embedded in a level set formulation to innovate a new cost function. Two dictionaries are build: A region-based feature dictionary and a voxel-wise dictionary. These dictionaries are learned, using the K-SVD method, from a public database of liver segmentation challenge (MICCAI-SLiver07). The learned dictionaries provide prior knowledge to the level set formulation. For the quantitative evaluation, the proposed method is evaluated using the testing data of MICCAI-SLiver07 database. The results are evaluated using different metric scores computed by the challenge organizers. The experimental results demonstrate the superiority of the proposed framework by achieving the highest segmentation accuracy (79.6\%) in comparison to the state-of-the-art methods.
	\end{abstract}

	\section{Introduction}
	\label{intro}
\begin{figure}[t]
\centering
\includegraphics[height=6.0cm]{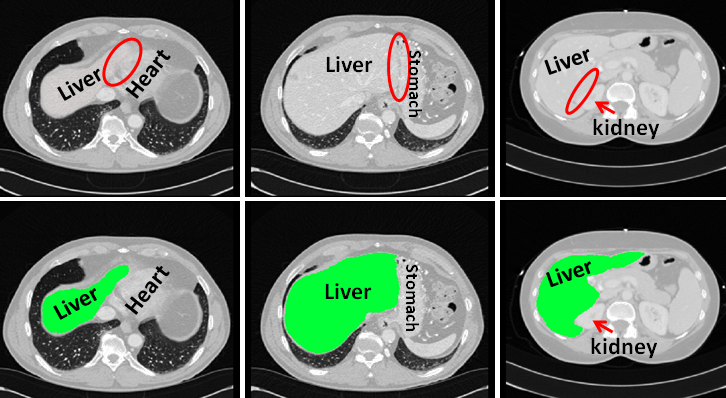}
   \caption{Overlapped boundaries in CT images of MICCAI-SLiver07 training data~\cite{Data2}. The red ellipses in the first row show the overlap of the liver-heart, liver-stomach, and liver-kidney boundaries respectively. The second row shows the liver ground truth segmentation.}
\label{fig:fig1}
\end{figure}

The liver is among the most common human organs to undergo invasive surgeries. In the case of liver tumors, an accurate 3D liver segmentation is important because a resection of the liver has to be carefully planned in order to preserve as much of the liver as possible~\cite{Campadelli}. Usually, computer tomography (CT) images are acquired for these purposes. 
The challenging aspects of liver segmentation in CT scan images can be summarized as follows: Firstly, the overlapping boundaries between the liver and surrounding organs such as heart, stomach, right kidney, and spleen, as illustrated in Fig.~\ref{fig:fig1}. Secondly, the large variability in liver shape, intensity distributions, and geometric properties from patient to patient, which make it difficult to describe the liver with model-based approaches~\cite{Campadelli}. Finally, liver segmentation, using a slice-by-slice approach, in 2D space is time consuming and gives inaccurate results. Therefore, for accurate segmentation results, volume segmentation methods in 3D space are more efficient~\cite{Campadelli}.

In recent years, a variety of methods have been proposed to segment the liver. In 2007, liver segmentation competitions from CT data were held in conjunction with MICCAI~\cite{Data2}. Between 2007-2014, more than 35 automatic methods were proposed and evaluated by the MICCAI-SLiver07 challenge organizers~\cite{Data2}. Most of them achieved segmentation accuracies between 60\% - 77\%. The best automatic method cited on the MICCAI-SLiver07 challenge~\cite{Data2} achieves 77.3\%~\cite{Kainmueller}. This method proposed a statistical deformable model with an intensity distribution model and modified the model segmentation results by a locally constrained free form deformation. Wimmer et al.~\cite{Wimmer} propose a probabilistic active shape model (ASM) using a level set method based on Parzen density estimation. This method is better than the ASM because it is able to capture more statistical information than the ASM. Linguraru et al.~\cite{Linguraru} propose a generic affine invariant shape parametrization using a graph-cut method for liver and liver tumor segmentation.

Moreover, there are many liver segmentation methods that were not involve in MICCAI-SLiver07 competition. Ling et al.~\cite{Ling} propose a hierarchical learning method to build a liver shape model. Wang et al.~\cite{Wang} propose a method to built a sparse representation of liver shape. This method needs a large training data to perfectly represent the shape prior. Zhang et al.~\cite{Zhang1} propose a voxel-wise-based segmentation method. This method considers only local image information to create the sparse representation of liver shape prior. Huang et al.~\cite{Huang1} propose an automatic method for liver segmentation. This method is based on probabilistic atlas. Tomoshige et al.~\cite{Tomoshige} propose a conditional statistical shape model (SSM) using non-contrast CT images. Huang et al.~\cite{Huang} propose an ASM based automatic liver segmentation method. This method is based on the ASM and boundary profile classification with free-form deformation.

Many approaches have shown that variational formulation is the most effective method to solve image segmentation problems but it needs well defined region boundaries or prior knowledge~\cite{Mumford}. In recent years, voxel-wise-based dictionary learning and sparse coding methods have been popular and useful tools in signal processing and machine learning. In image segmentation, voxel-wise-based approaches, although they can be useful, achieve limited success because of the non-linear distribution of the image data such as the overlaps between region boundaries in medical images~\cite{Al-Shaikhli}. Therefore, a perfect segmentation needs a non-linear separation of the image data. One useful solution to this problem is the mapping of the image data from the data space to the feature space. Such a mapping allows for a linear separation of the overlapping regions.

To solve the above mentioned limitations of the variational formulation and voxel-wise dictionary learning methods, in this paper, we present a novel fully automatic liver segmentation framework via a level set formulation using dictionary learning and sparse coding with shape prior. The proposed level set energy equation consists of data and regularization terms. These terms have been integrated into the implicit framework in a novel fashion. The data term consists of the sparse representation of global image features, and the shape prior, which represents the voxel-wise sparse representation (local image information) of liver shapes. The regularization term is modified to include the shape prior. The main \textbf{contribution} of this paper is the embedding of the global and local sparse representation of image information into the data term to integrate a novel cost function of the image data. Two dictionaries are built: one using global image features (texture, CT number, and volume properties) and one using voxel-wise-based learning. The remaining sections are organized as follows. The proposed framework is described in detail in Sec.~\ref{sec:method}. In Sec.~\ref{sec:expRes}, experimental results are presented and discussed. Finally, this work is concluded in Sec.~\ref{sec:conc}.
\section{Method}
\label{sec:method}

\subsection{Feature Extraction}
\label{ssec:FE}

Sets of features are considered to learn the dictionaries: texture, volume properties, and Hounsfield scale (HU). Hounsfield scale (HU) or called CT-number of the liver varies between $(+40 - +60 HU)$. A set of texture features is considered in our method (entropy, energy, contrast, homogeneity, sum mean, variance correlation, maximum probability, inverse difference moment, cluster tendency). These features have been computed using the gray level co-occurrence matrix (GLCM) for four different offsets $(0^\circ,\:45^\circ,\: 90^\circ,$ and $\;135^\circ)$~\cite{Haralick}. 
Volume properties are important and they refer to the topological features of the liver. We use the volume, surface area, Euler number, major axis length, and minor axis length~\cite{Schladitz}. The total number of features used, in our method, is 42 features.   
\subsection{Dictionary Learning}
\label{ssec:DL}

In our framework, MICCAI-SLiver07 training data~\cite{Data2} (20 patients CT scan image data) are used for learning. Using the ground truth segmentation, each image data of the training set is considered as two labels (liver/non-liver). A dictionary is built for each label. Let $c=1,2$ be a number of the labels in the image data, $N_c$ are the training set of each label. The goal is to build a dictionary matrix, for each label, that has a perfect representation of $N_c$.

Let $D_c$ be a dictionary $n\times z\times K_c$ matrix $D_c=(d_1, d_2,...,$ $d_{K_c})$, which consists of $K_c$ atoms (columns), $\{d_i  \in R^{n\times z}: i=1, 2,..., K_c\}$ and each atom represents the key features of $Y_c$, where $(K_c~\ll~N_c)$. $Y_c=(y_1, y_2,..., y_{N_c})$ is a $n\times z \times N_c$ matrix which consists of feature matrices $\{y_i \in R^{n\times z}: i=1, 2,..., N_c\}$ of $N_c$ data samples with dimension $n\times z$. $n=42$ is the number of features used to train the dictionaries, and $z=160$.
The sparse representation $A_c=(a_1, a_2,..., a_{N_c}) \in R^{K_c \times N_c\times z}$ is computed s.t. $y_i=D_ca_i$ and $\Vert a_i\Vert_0 << K_c, i=1,..., N_c$. In such a way that each feature matrix in $Y_c$ is represented by linear combination of a few atoms in the dictionary according to the non-zero elements in $A_c$. The problem can be formulated as the following minimization:
\begin{equation}
\label{eq:eq2}
\arg\min_{D_c,A_c}\Vert Y_c-D_cA_c \Vert_F^2\;\;
s.t.\;\;\forall 1\leq i \leq N_c , \; \Vert a_i \Vert_0 \ll K_c
\end{equation}
To solve Eq.~(\ref{eq:eq2}), we propose the use of K-SVD method~\cite{Aharon}. This method is robust to solve the problem in Eq.~(\ref{eq:eq2}) by iterating $K_c$ times of singular value decomposition (SVD) for the dictionary update step. For sparse coding step, we use the orthogonal matching pursuit (OMP).
\subsection{The Proposed Level-Set Framework}
\label{ssec:lsf}
The proposed energy equation is based on the piece-wise constant of the Mumford-Shah variational model~\cite{Mumford}:
\begin{equation}
\label{eq:eq1}
F(M,B)=\underbrace{\int_{R}(I-M)^2 dx}\limits_{data~term} ~+ \underbrace{\lambda r(B)}\limits_{regularization~term}
\end{equation}
where $M:R\rightarrow \mathcal{R}^2$ is a constant approximation of the observed image $I$. $r(B)$ is the length of the boundary of $M$.

Due to the difficulties illustrated in Fig.~\ref{fig:fig1}, in this paper, we consider two prior knowledge to formulate the proposed level-set formulation: the global image features, and the local image information, which represents the shape prior. 

We consider two steps for automatic initial liver localization in the target image. The first step is thresholding the target image. This step is based on that the gray level range of the liver is between 125-155. Second, the CT number of the liver is between $(+40 - +60 HU)$. The second step is based on the criteria that the liver is the biggest organ in the abdominal cavity and it lays on the right side. In this step, the image data (after thresholding) is divided into left and right side images. The centroid of the segmented volume in the right side image is computed. Then, from the centroid, $16\times 16\times 16$ bounding box is computed to represent the initial level set. 
\subsubsection{Global (Region-based) Image Features}
\label{ssec:gi}
In this subsection, we describe the proposed method using the learned dictionary $D_c$ with the level set formulation. Starting from the Mumford variational model~\cite{Mumford}, the data term ($\int_R (I-M)^2$) in Eq.~(\ref{eq:eq1}) can be considered as a k-means clustering problem if the regularization term is ignored. Therefore, this term can be reformulated as $\ell_2$ norm, assuming that the $\ell_2$ is a generalization of k-means clustering problem:
\begin{equation}
E_1(A_c)=P (\min_{A_c} \parallel V_f-D_cA_c\parallel_2^2)  + \lambda|b|
\label{eq:eq6c}
\end{equation}
\begin{equation}
\label{eq:eq4.5}
P(l)= \left\{ \begin{array}{ll} \geq 0 \;\; &\mbox{if \; $l \in V_f$} \\ < 0 \;\; & \mbox{if \; $l \in V_f^{com}$}
\end{array}\right.
\end{equation}
\begin{equation}
\label{eq:eq4.5z}
H(P)= \left\{ \begin{array}{ll} 1 \;\; &\mbox{if \; $P \geq 0$} \\ 0 \;\; & \mbox{if \; $P < 0$}
\end{array}\right.
\end{equation}
where $E$ is the energy function. $V_f$ is a matrix of feature samples of the liver, with the size $n\times z$, in the target image. $\lambda >0$ is the regularization parameter and $b$ is the contour length. $l$ is the label of voxels in $V_f$, and $V_f^{com}$ is the complement of $V_f$. $P(l)$ represents the descriptor of the liver volume in the target image. $H(P)$ is a Heaviside function.
\subsubsection{Local (voxel-wise) Image Information (Shape Prior)}
\label{ssec:li} 
Since the gray-values in all organs are highly similar, the global image information is infeasible to achieve perfect result. Therefore, we add the local image information as prior knowledge. The local image information is represented by the shape prior, which represents the sparse representation (voxel-wise) of the liver shape.

Using the ground truth segmentation of the training data~\cite{Data2}, we build the voxel-wise sparse representation of $N$ liver shapes. From the image data, patches are extracted. Each patch, with a size $16\times 16\times z$, is concatenated in a matrix with size $n_s\times z$, where $n_s=256$ and $z=160$. Thus, $D_s=[d_1, d_2,\dots, d_{K_s}]\in R^{n_s,z,K_s}$, $\{d_i \in R^{n_s\times z}: i=1, 2,..., K_s\}$, and $K_s$ is the total number of columns in $D_s$. In such a way that each sample matrix in $\{Y_{s_i} \in R^{n_s\times z},~i=1, 2,..., N\}$ is represented by linear combination of a few atoms in the dictionary $D_s$ according to the non-zero elements in the sparse representation $\alpha$. $\alpha=(\alpha_1, \alpha_2,..., \alpha_{N}) \in R^{K_s \times N}$ is computed s.t. $Y_{s_i}=D_s\alpha_i$ and $\Vert \alpha_i\Vert_0 << K_s, i=1,..., N$. 

Let $I$ be a test image data and let $V_t$ be an initial liver volume in $I$, i.e. initially detected VOI as explained in the beginning of Section~\ref{ssec:lsf}. To find the desired shape, the patches are extracted and concatenated in a matrix as explained above. Then, we solve the following minimization problem:
\begin{equation}
\min_\alpha \parallel \alpha\parallel_0 ~~~ s.~t. ~~~ \parallel V_t-D_s \alpha\parallel_2 \leq \epsilon
\label{eq:eq4c}
\end{equation}
One possible solution of Eq.~\ref{eq:eq4c} is the replacement of the $\ell_0$ norm with $\ell_1$ norm~\cite{Chen}: 
\begin{equation}
E_{shape}(\alpha)=\min_\alpha\parallel V_t-D_s \alpha\parallel_2^2 + \lambda\parallel \alpha\parallel_1
\label{eq:eq5c}
\end{equation}
In Eq.~(\ref{eq:eq5c}), the first term is the data term. The second term is the regularization term of the shape prior. If the solution of Eq.~(\ref{eq:eq5c}) exists, the desired shape is obtained, i.e. $\parallel \alpha\parallel_1$ should has only one non-zero entrance which represents the desired shape prior.

Combining Eq.s (~\ref{eq:eq6c}) and~(\ref{eq:eq5c}), the overall proposed energy equation can be formulated as follows:
\begin{equation}
\label{eq:eq12c}
E_{total}(A_c,\alpha)= \underbrace{H~G}\limits_{Foreground}+\underbrace{H~G^c}\limits_{Background}+\underbrace{\lambda(|b| + \parallel \alpha\parallel_1)} \limits_{Regularization}
\end{equation}
where $G$ is:
\begin{equation}
G=(\min_{A_c} \parallel D_cA_c-V_f\parallel_2^2)(\min_\alpha\parallel D_s\alpha-V_t\parallel_2^2),\;\;\; G^c=1-G
\end{equation}
\subsubsection{Level-Set Optimization}
\label{sssec:opt}
This subsection presents the optimization of the proposed energy function (Eq.~\eqref{eq:eq12c}). The main task, to get this minimization, is by finding the best matching between the sparse representation of the target image data with learned dictionaries $D_c$ and $D_s$. 

Once the $\ell_1$~norm is applied to the constraint term, Eq.~(\ref{eq:eq12c}) become difficult to solve. Here, we adopt the concept of iteratively re-weighted (IR) algorithm to handle this challenge. We adopt the general idea of IR algorithm by reformulating the minimization problem in Eq.~(\ref{eq:eq12c}) to the weighted mean square error (MSE) at the $t^{th}$ iteration, given by: At iteration $(t)$, we consider $W^t$ as the weight of the volume of interest (VOI) at $(t-1)$ iteration, i.e. $W_1^t=H\min_{A_c}\parallel V_f-D_cA_c^{t-1}\parallel_2^2$ and $W_2^t=H\min_{\alpha_c}\parallel V_t-D_s\alpha^{t-1}\parallel_2^2$:
\begin{equation}
A_c^t \leftarrow A_c^{t-1}
\label{eq:eq13c}
\end{equation}
\begin{equation}
\alpha^t  \leftarrow \alpha^{t-1}
\end{equation}
In all experiments, 
$\lambda$ is set to 0.7.
\section{Experimental Results and Discussion}
\label{sec:expRes}

In this paper, several experiments have been conducted on diverse medical images to explore the advantages of the proposed method. We use a public CT-scan medical image database, namely, the segmentation of the liver 2007 (MICCAI-SLiver07) database~\cite{Data2} (20 patients training data and 10 patients testing data). The MICCAI-SLiver07 database is publicly available through the MICCAI 2007 Segmentation of the Liver challenge~\cite{Data2}. This image database provides the ground truth segmentation of the training data (manual expert annotations), and have a slice resolution of 512$\times$512 voxel and 64 to 502 slices. The inter-slice space varies between 0.7 mm and 5 mm and the intra-slice space varies between 0.56 mm$\times$0.56 mm and 0.86 mm$\times$0.86 mm. 

\begin{figure}[t]
\centering
\includegraphics[height=7.0cm]{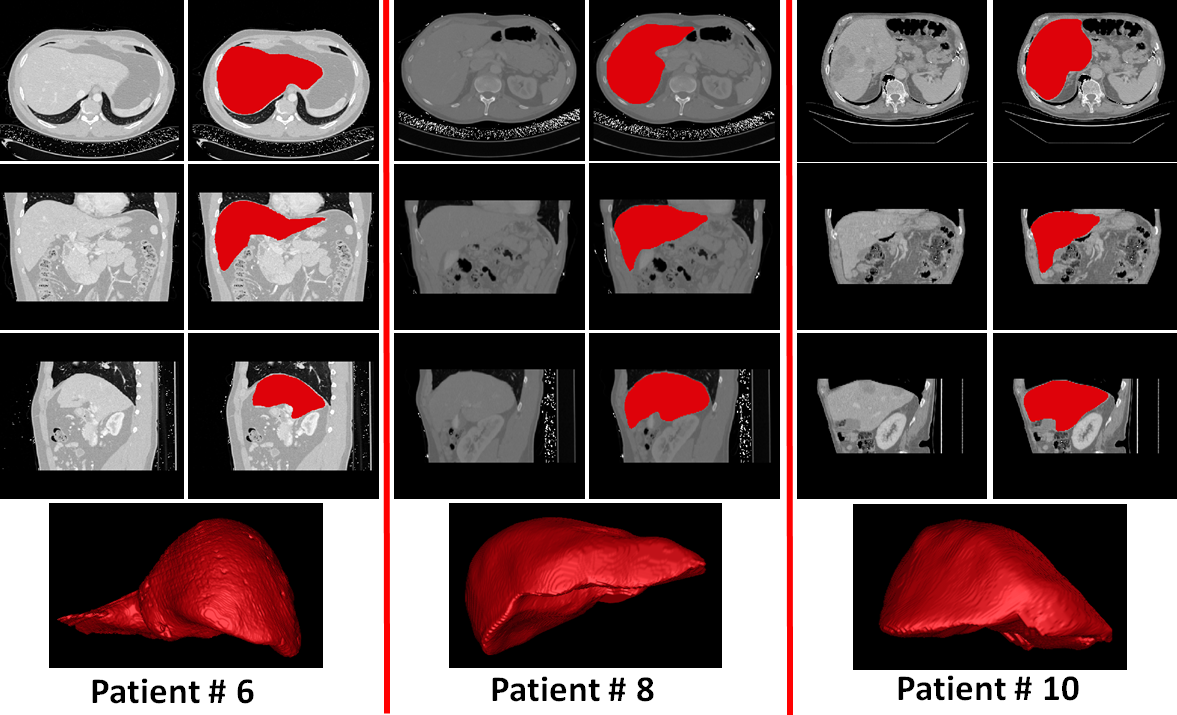}
   \caption{Three liver segmentation examples using the MICCAI-SLiver07 testing data~\cite{Data2}. In each example, the first column shows the original image data in coronal, sagittal, and axial planes. The second column represents the segmentation results of these sections. The fourth row shows the 3D liver segmentation of our method.}
\label{fig:fig5}
\end{figure}

Figure~\ref{fig:fig5} shows three examples of our results of the MICCAI-SLiver07 testing data~\cite{Data2}. In each example, the first column is the original image data. The second column shows the segmentation results. The fourth row represents the 3D liver segmentation results. 
The examples given in Figs.~\ref{fig:fig5} have a different contrast. The boundary ambiguity between the liver and surrounding organs is noted in all examples. The results show the ability of our method to solve this segmentation challenge by considering the global and local image information in the level set formulation.

The performance of the proposed method has been evaluated using five metrics proposed by the MICCAI-SLiver07 challenge organizers~\cite{Data2}. First, the overlap error is defined as VOE$=100(1-(|H\cap T|/|H\cup T|))$ and is given in percent. Second, the relative volume difference is defined as VD$=100(|H|-|T|/|T|)$ and is given in percent. It gives an indication whether the results are over- or under-segmentation. Third, the average symmetric surface Distance (AvgD) is the average distance of all surface voxels distances between each surface voxel in $H$ and the closest surface voxel in $T$. Forth, the root mean square symmetric surface distance (RMSD) is based on the surface distance and is given in millimeters. Fifth, the maximum symmetric surface distance (MaxD) represents the maximum differences between both sets of surface voxels in $H$ and $T$ and is given in millimeters. It also known as the Hausdorff distance. Assuming that $H$ and $T$ are two sets of voxels of the segmentation results and ground truth segmentation respectively.

The results of the testing data were sent to the MICCAI-SLiver07 challenge organizers and the evaluation is obtained by the organizers according to the conditions of the challenge~\cite{Data2}. 
Table~\ref{table:tab1} shows the results of 10 patients of SLiver07 testing data~\cite{Data2}. Table~\ref{table:tab1} illustrates five different metrics (VOE, VD, AvgD, RMSD, and MaxD) with the average value (Avg) of all of these metrics. 
\setlength{\tabcolsep}{0.5pt}
\begin{table}[!t]
\centering
\caption{Our average results of MICCAI-SLiver07 testing data~\cite{Data2} using evaluation metrics~\cite{Data2}. The maximum scores (Scr) is 100. The results for each metric report the mean over all images, together with mean scores. All scores are averaged to a final score over all images. Note: This results are obtained by MICCAI-SLiver07 challenge organizers and they are cited on~\url{http://sliver07.org/showresult.php?rank=12&submission=2014-09-20-2136}.}
\label{table:tab1}
\begin{tabular}{|l||c|l||c|l||c|l||c|l||c|l||c|l||c|l||c|l||c|l||c|l||c|l|}
\hline
\textbf{Method} & \textbf{VOE} & \textbf{Scr} & \textbf{VD} &\textbf{Scr} & \textbf{AvgD} & \textbf{Scr} & \textbf{RMSD} & \textbf{Scr} & \textbf{MaxD} & \textbf{Scr} & \textbf{Total} \\
 & \textbf{[\%]} & & \textbf{[\%]} &  & \textbf{[mm]} &  & \textbf{[mm]} &  & \textbf{[mm]} & & \textbf{Scr} \\
\hline
1 & 6.74 & 73.7 & 2.45 & 87.0 & 0.94 & 76.4 & 1.48 & 79.5 & 14.98 & 80.3 & 79.4 \\
2 & 7.49 & 70.7 & 3.35 & 82.2 & 1.10 & 72.5 & 2.18 & 69.7 & 25.61 & 66.3 & 72.3 \\
3 & \textbf{5.30} & \textbf{79.3} & 0.92 & 95.1 & 1.01 & 74.7 & 1.57 & 78.2 & 22.21 & 70.8 & 79.6 \\
4 & 6.32 & 75.3 & -0.81 & 95.7 & 0.92 & 76.9 & 1.55 & 78.4 & 13.98 & 81.6 & 81.6 \\
5 & 6.20 & 75.8 & 1.62 & 91.4 & 1.02 & 74.5 & 1.90 & 73.7 & 19.40 & 74.5 & 78.0 \\
6 & 6.55 & 74.4 & \textbf{-0.15} & \textbf{99.2} & 0.99 & 75.3 & 1.51 & 79.0 & 13.14 & 82.7 & 82.1 \\
7 & 6.30 & 75.4 & 3.70 & 80.3 & 0.89 & 77.8 & \textbf{1.33} & \textbf{81.6} & \textbf{10.34} & \textbf{86.4} & 80.3 \\
8 & 6.17 & 75.9 & 3.39 & 82.0 & 0.97 & 75.8 & 1.51 & 79.0 & 11.81 & 84.5 & 79.4 \\
9 & 7.53 & 70.6 & 1.92 & 89.8 & 0.88 & 78.0 & 1.36 & 81.2 & 17.09 & 77.5 & 79.4 \\
10 & 5.78 & 77.4 & -1.05 & 94.4 & \textbf{0.81} & \textbf{79.8} & 1.40 & 80.6 & 10.61 & 86.0 & \textbf{83.7} \\
\hline
\rowcolor{MRed}Avg. & 6.44 & 74.9 & 1.53 & 89.7 & 0.95 & 76.3 & 1.58 & 78.1 & 15.92 & 79.1 & 79.6 \\
\hline
\end{tabular}
\end{table}
\setlength{\tabcolsep}{0.5pt}

The proposed method achieves the highest rank (segmentation accuracy) among 35 automatic methods submitted in~\cite{Data2a}. The results of our method have also been published online~\cite{Data2a}. In MICCAI-SLiver07 challenge, some automatic methods, such as the methods presented in~\cite{Kainmueller,Linguraru}, used large training data and obtained their results using the training data of 112 and 92 training image respectively instead of the 20 patients.

The evaluation of the MICCAI-SLiver07 challenge organizers~\cite{Data2} has become a reference for liver segmentation enabling efficient and precise comparisons~\cite{Data2}. Thus, we compare the proposed method to the best automatic methods~\cite{Kainmueller,Wimmer,Linguraru}, which were evaluated by challenge organizers, as shown in Table~\ref{table:tab2}. 
\setlength{\tabcolsep}{2.2pt}
\begin{table}[!t]
\centering
\caption{Our results compared to the methods~\cite{Kainmueller,Wimmer,Linguraru}. The maximum scores (Scr) is 100\%. The results for each metric report the mean over all testing data, with mean scores.}
\label{table:tab2}
\begin{tabular}{|l||c|l||c|l||c|l||c||l||c|}
\hline \centering
\textbf{Alg.} & \textbf{VOE} & \textbf{Scr} & \textbf{AvgD} & \textbf{Scr} & \textbf{RMSD} & \textbf{Scr} & \textbf{\#~Vol.} & \textbf{\#~Train.} & \textbf{Average} \\
 & \textbf{[\%]} & \textbf{[\%]} & \textbf{[mm]} & \textbf{[\%]} & \textbf{[mm]} & \textbf{[\%]} &  \textbf{Tested} & \textbf{Vol.} & \textbf{Scr~[\%]} \\
\hline
\rowcolor{MRed}
Ours & 6.44 & 74.9 & \textbf{0.95} & \textbf{76.3} & \textbf{1.58} & \textbf{78.1} & 10 & 20 & \textbf{79.6} \\\hline
Alg.\cite{Kainmueller} & \textbf{6.09} & \textbf{76.2} & \textbf{0.95} & \textbf{76.3} & 1.87 & 74.0 & 10 & 112 & 77.3 \\\hline
Alg.\cite{Wimmer} & 6.47 & 74.7 & 1.02 & 74.5 & 2.00 & 72.3 & 10 & 20 & 76.8 \\\hline
Alg.\cite{Linguraru} & 6.37 & 75.1 & 1.00 & 74.9 & 1.92 & 73.4 & 10 & 92 & 76.2 \\\hline
\end{tabular}
\end{table}
\setlength{\tabcolsep}{2.2pt}
\section{Conclusion}
\label{sec:conc}

This paper presents a fully automatic 3D liver segmentation method based on level set formulation using dictionary learning with shape prior. We integrate the data and regularization terms of the level set energy equation in a novel fashion. Two dictionaries are used in our framework: the feature dictionary, which represents the global image features, and the voxel-wise dictionary, which represents the shape prior (local image information). The proposed method has been tested using MICCAI-SLiver07 database~\cite{Data2}. The evaluation results of the MICCAI-SLiver07 testing data are computed by the MICCAI-SLiver07 challenge organizers. The experimental results show that our method achieves the best segmentation accuracy (79.6\%) among more than 35 automatic segmentation methods cited on MICCAI-SLiver07 challenge website~\cite{Data2a}.
%
\label{references}

\end{document}